\documentclass{jmlr} % use larger type; default would be 10pt
\usepackage{microtype}
\usepackage[utf8]{inputenc} % set input encoding (not needed with XeLaTeX)
\usepackage{graphicx} % support the \includegraphics command and options

\usepackage{authblk}
\usepackage{hyperref}
\newcommand{\reg}{\textsuperscript{\textregistered}}
\newcommand{\naive}{na\"{\i}ve}
\newcommand{\vs}{vs.\ }
\title[Evaluation Toolkit for Causal Inference]{An Evaluation Toolkit to Guide Model Selection and Cohort Definition in Causal Inference}
\author[1,*]{Yishai Shimoni}
\author[1]{ Ehud Karavani}
\author[1]{Sivan Ravid}
\author[1]{Peter Bak}
\author[2]{Tan Hung Ng}
\author[3]{Sharon Hensley Alford}
\author[3]{Denise Meade}
\author[1]{Yaara Goldschmidt}

\affil[1]{IBM Research - Haifa, Israel}
\affil[2]{IBM Watson Health, San Jose, CA}
\affil[3]{IBM Watson Health, Cambridge, MA}
\affil[*]{yishais@il.ibm.com}

\begin{document}
    \maketitle

    \begin{abstract}
        Real world observational data, together with causal inference, 
        allow the estimation of causal effects when randomized controlled trials are not available.
        To be accepted into practice, such predictive models must be validated for the dataset at hand, 
        and thus require a comprehensive evaluation toolkit, as introduced here.
        Since effect estimation cannot be evaluated directly, we turn to evaluating the various observable properties of causal inference,
        namely the observed outcome and treatment assignment.

        We developed a toolkit that expands established machine learning evaluation methods and adds several causal-specific ones.
        Evaluations can be applied in cross-validation, in a train-test scheme, or on the training data.
        Multiple causal inference methods are implemented within the toolkit in a way that allows modular use of the underlying machine learning models.
        Thus, the toolkit is agnostic to the machine learning model that is used.
        We showcase our approach using a rheumatoid arthritis cohort (consisting of $\approx 120$K patients)
        extracted from the IBM MarketScan\reg Research Database.

        We introduce an iterative pipeline of data definition, model definition, and model evaluation.
        Using this pipeline, we demonstrate how each of the evaluation components helps drive model selection 
        and refinement of data extraction criteria in a way
        that provides more reproducible results and ensures that the causal question is answerable with available data.
        Furthermore, we show how the evaluation toolkit can be used to ensure that performance is maintained when applied to subsets of the data,
        thus allowing exploration of questions that move towards personalized medicine.
        Overall, the iterative pipeline and the evaluation toolkit provide a guideline for best practices in model selection
        and data extraction for causal inference in healthcare 
        and pave the way for the application of more reliable models in real-world scenarios.
    \end{abstract}

    \section{Introduction}
    Causal inference analysis leverages observational data to estimate the effect of an intervention on some outcome \citep{pearl2009causal}.
    In healthcare, it uses observed outcomes and observed treatment assignments to help predict how those outcomes might have changed,
    on average, had each individual been assigned some treatment of interest (which may differ from the one they were observed to receive).
    When possible, such treatment questions are answered on a sample of the population level using randomized controlled trials (RCTs),
    in which individuals are randomly assigned an intervention.
    However, causal inference analyses using real-world evidence (RWE) data are essential when RCTs are infeasible
    due to time, expense, or ethics \citep{deaton2018understanding}.

    In recent years large RWE data has become more readily available,
    allowing a resurgence of interest in causal inference \citep{Pearl2018bookofwhy, hernan2018data},
    leading to its application in decision making within healthcare as well as in other industries
    \citep{pearl2009causal, hernan2006estimating, morgan2015counterfactuals, gassen2014causal}.
    It was also noticed recently that in order to determine the best therapy for an individual (in the context of personalized medicine)
    accurate and reliable causal inference models that were trained on large datasets are required \citep{shalit2016estimating}.
    Thus, RWE combined with causal inference are essential for decision making in healthcare at all scopes.

    While RWE allows asking health-related questions that cannot be addressed by RCTs,
    observational data are known to contain biases related to treatment assignment \citep{cochran1973controlling}.
    Therefore, machine learning models are often used within causal inference methods to correct for biases 
    and help estimate causal effects \citep{peters2017elements}.

    Another problem concerning RWE data, stated by "the fundamental problem of causal inference"
    \citep{holland1986statistics}, is that the \emph{effect} of a treatment (unlike the outcome) is never observable
    and requires estimation of what would have been the outcome had each person received a treatment
    other than the one that was observed (counterfactual outcomes).
    Consequently, effect estimation cannot be evaluated using standard machine learning evaluation approaches.

    One common solution is to evaluate causal inference performance using a synthetic benchmarking dataset
    \citep{guyon2011causality, mccann2010causality, shimoni2018benchmarking, mooij2016distinguishing, dorie2017automated}.
    This allows all potential outcomes to be simulated under each possible treatment rendering the effect of each treatment observable,
    thereby allowing estimation of the method's performance.
    However, as the no-free-lunch theorem for optimization \citep{wolpert1997no} implies,
    the generalizability of the performance to real data cannot be assumed.
    It is therefore necessary to evaluate the performance of each method on each individual dataset \citep{johnson2009good}.
    Moreover, the effect of the treatment is often estimated using the same data on which the model was trained \citep{hernan2018causal}.
    While these two practices (benchmarking and effect estimation on the training data) 
    have proved to be useful in finding good causal inference methods,
    they can detract from the generalizability of both the models and their conclusions, which can lead to barriers in model and method adoption.

    While there have been several efforts to address this issue in the past \citep{van2007super},
    this problem has been recognized recently by the observational health data science and informatics (OHDSI) consortium 
    \citep{hripcsak2015observational},
    as a critical and timely one that requires additional research.
    The consortium provides R libraries for training causal inference methods and several evaluations of model performance 
    \citep{vashisht2018association, reps2018design, schuemie2018cohortmethod}.
    We provide here an evaluation toolkit, written in Python, that includes both similar evaluations and several novel ones,
    and furthermore extends the availability of such tools to the popular Python programming language.
    Moreover, our toolkit focuses more on the causal-specific aspect of model evaluation.
    Detailed explanations of the evaluations and their application to a specific example are provided below.
    
    \subsection*{Clinical Relevance}
        Most decisions in healthcare involve asking how some clinical, safety, cost, or utilization outcome might change if things are done differently.
        Causal inference analysis is designed to answer such questions in a way that emulates a random control trial from observational data.
        However, a combination of two issues have hindered the acceptance of such analyses into decision making in healthcare.
        The first is that such methods are eventually based on machine learning models, and thus require evaluations.
        The second is that the predictions of causal inference analysis are inherently impossible to evaluate directly.
        To address this we developed a method that evaluates the machine learning part of the causal inference analysis 
        and can provide good indication of the performance of the method, 
        and can therefore allow introduction of such methods into decision making.
        We provide here an example for the analysis of the risk of anemia in rheumatoid arthritis patients when treated with TNF-alpha blockers,
        compared to other biological treatments.
        The example shows that using the Marketscan\reg data we can create reliable causal inference models that can predict
        the risk of anemia on the whole population, as well as within sub-populations of interest.
        
    \subsection*{Technical Significance}
        Causal inference analysis is a field that is gaining renewed interest with the increase in availability of large databases,
        and allows emulating randomized controlled trials.
        However, the fact that neither the counter-factual outcome (the outcome that would have been observed if given another treatment)
        nor the propensity to treat are observed leads to the fact that causal inference predictions cannot be directly evaluated.
        This creates a barrier in the acceptance of such models into practice.
        We propose here to evaluate the performance of the underlying machine learning models on properties that can be observed and evaluated,
        namely the observed outcomes, and the observed treatment assignments.
        We adjust known machine learning evaluation tools to fit the context of causal inference.
        While these evaluations cannot prove that the models are predicting effect well, 
        they can detect bad performance and provide a level of confidence in the provided results.
        We present here a toolkit of evaluation tools, aimed at identifying various causal inference related problems, 
        and show through an example how these evaluations can help 
        adjust the models, the analyses, and the causal questions that can be answered on a given dataset.

    \section{Methods}
    Most approaches to estimate the causal effect of treatment on an outcome can be categorized as either balancing the treatment groups
    (e.g.~using propensity score for weighting or matching)  \citep{horvitz1952generalization} and/or modeling the
    outcome response surface directly \citep{gelman2006data}.
    Both cases predict some value based on the individual patient characteristics,
    and therefore require employing an underlying machine learning model,
    which in turn requires performance evaluation.

    Let us first consider outcome prediction models.
    As mentioned above, the effect of an intervention can never be observed,
    and therefore effect estimation performance is impossible to fully evaluate.
    Nevertheless, for a causal inference model to be correct it must at least be correct on the predictions of the factual (i.e., observed) outcomes,
    and the performance for this task can be evaluated.
    Thus, re-formulating the performance evaluation in such a way allows applying established evaluation methods
    from statistics and machine learning to the field of causal inference.
    Specifically, under the assumptions of causal inference we can assume that
    $\Pr[Y^a|X]=Pr[Y|A=a, X]$,
    where $Y, A, X$ represent the outcome, the treatment, and the covariates, respectively,
    and the notation is as defined in \nameref{appendix_A}.

    For propensity models, while the propensity is not directly observed, the treatment assignment is observed,
    and therefore average propensity $\Pr[A|X]$ can be estimated.
    We can therefore evaluate how well the model predicts the propensity itself,
    and additionally, how well the model predicts the observed assignment directly.
    For weighting models, we can estimate how well they allow emulating RCTs from the observed data by removing treatment bias.
    In this context, it is important to note that matching algorithms can be considered as integer weighting algorithms,
    and therefore any further reference to weighting algorithms applies to matching as well.

    \begin{figure}
        \begin{center}
            \includegraphics[width=\columnwidth]{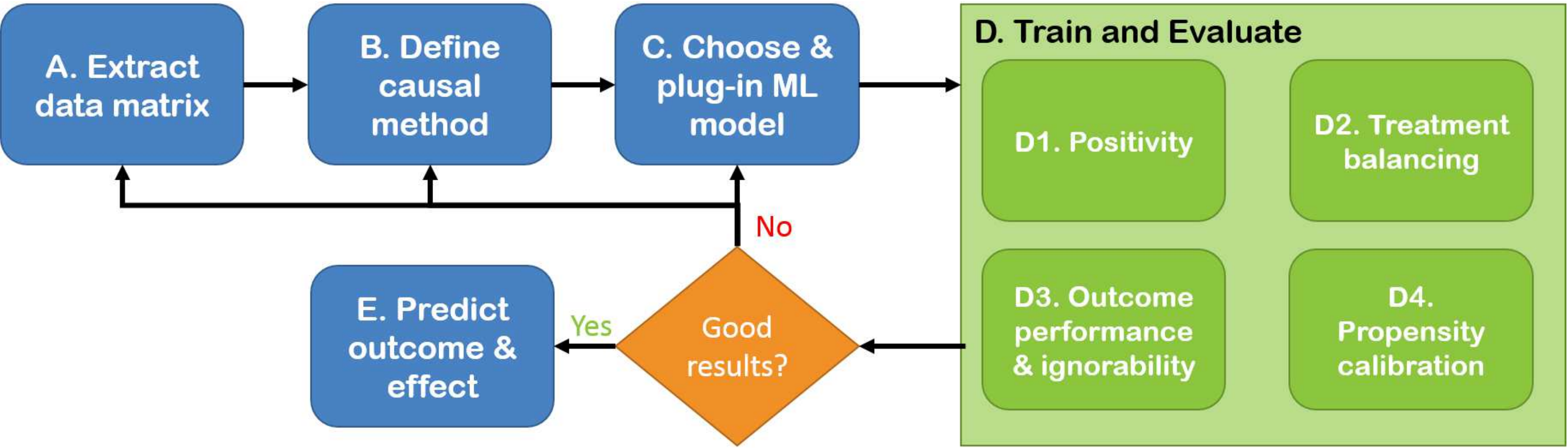}
            \caption{A schematics of the pipeline to guide model selection and cohort definition in causal inference.
            The pipeline involves an iterative process, in which a) the causal inference defined and a data matrix is extracted;
            b) the causal methods is chosen;
            c) the underlying machine learning models are chosen;
            d) the model performance is evaluated.
            If the models perform well then causal inference prediction can be drawn to estimate outcome and effect.
            Otherwise, the process need to be reiterated following some modifications in steps a-c}
            \label{fig:eval_pipeline}
        \end{center}
    \end{figure}

    Using these insights, we implemented a toolkit of methods to evaluate the performance of the underlying machine-learning models,
    whether they predict propensity, sample weights, or outcome, and suggest an iterative flow for applying these methods.
    This flow consists of repeating the following steps, as depicted in Figure \ref{fig:eval_pipeline}:
    a) Define the causal question and extract the data accordingly;
    b) Choose the causal inference method;
    c) Choose the machine learning model to be used within the causal inference method;
    d) Train the model in cross-validation form and evaluate the performance on both training and held-out set;
    e) If the performance is not satisfactory, repeat the pipeline, adjusting steps a-c.
    If the performance is satisfactory, apply the causal inference to extract the desired insights.

    Whether using a propensity model, a weight model,
    or an outcome model, our evaluations help determine the best underlying machine-learning models for any specific prediction task
    by providing multiple metrics and summary visualizations.
    These evaluations are derived from standard machine learning evaluations,
    but many are tailored specifically for causal inference evaluation, as described individually below.

    In the results section we apply our evaluation toolkit method to an example of a rheumatoid arthritis (RA)
    patient cohort extracted from the IBM Marketscan\reg Research Databases.
    In this example we applied a doubly-robust method that combines both
    a propensity model and an outcome model
    (using the inverse of the propensity as an additional feature for the outcome model), as an example of its use.
    We stress, however, that the toolkit can be used for any causal method
    that provides predictions of either propensity scores, weights, or counterfactual outcomes.
    Similarly, in the example we utilized simple, out of the box, well known causal methods and machine learning models
    solely for the purpose of exemplifying the use of the evaluation toolkit.
    Any other causal method or machine learning model can be used instead.

    \subsection{Metrics and Plots}
    We developed our causal inference evaluation toolkit in Python 3.6, with an interface compatible with scikit-learn \citep{scikit-learn}.
    In the course of this work we adapted statistical and machine learning approaches and developed several novel visualizations.
    Our toolkit includes applying \emph{receiver operator characteristic} (ROC) curves showing the true positive rate \vs false positive rates;
    \emph{balancing plots} displaying the absolute standardized mean difference between the treated and untreated for each covariate \citep{love2002displaying};
    \emph{calibration plots} to check accuracy \citep{zadrozny2002transforming};
    plots comparing the \emph{distribution of predicted propensity} in each treatment group;
    \emph{accuracy plots} showing the predicted outcome \vs the observed outcome;
    and \emph{counter-factual scatter plots} comparing the predicted outcome in each treatment group.
    Since most of the evaluations utilize tools that are standard in machine learning,
    the technical details for how these plots are created can be found in \nameref{appendix_A}.

    \subsection{Data Source}
    The example data used here was extracted from two IBM MarketScan\reg Research Databases, Commercial Claims and Medicare Supplemental,
    which contain individual-level, de-identified, healthcare claims information from employers, health plans, hospitals,
    as well as Medicare supplemental programs, for the period of January 1st, 2010 to April 30th, 2017.
    Specifically:
    a) IBM MarketScan\reg Commercial Database contains health insurance claims across the continuum of care
    (e.g.~inpatient, outpatient, and outpatient pharmacy)
    as well as enrollment data from large employers and health plans across the United States
    which provide private healthcare coverage for more than 100 million employees, their spouses, and dependents;
    and b) IBM MarketScan\reg Medicare Supplemental Database is created for Medicare-eligible retirees with
    employer-sponsored Medicare Supplemental plans.

    The data for individual patients is integrated from all care providers, maintaining all healthcare
    utilization and cost record connections at the patient level.

    \section{Results}
    Instead of listing all the evaluation options in our evaluation toolkit we use a specific cohort and a specific causal question,
    as described below, as an example throughout this section.
    Using this example, we demonstrate the use of the evaluation toolkit in a pipeline to guide both the model selection
    and the data definition as described in Figure~\ref{fig:eval_pipeline}.
    Using this pipeline, we highlight both bad practices and good ones.
    We do not focus on the details of the clinical question that is addressed here or on the clinical implications of the results.

    Step A in Figure~\ref{fig:eval_pipeline} requires defining the causal question and extracting the data.
    To demonstrate the use of our toolkit, consider estimating the effect of treatment with a TNF-alpha antagonist (TNFA)
    compared to treatment with other biologic or conventional disease modifying antirheumatic drugs (DMARD),
    for recently-diagnosed rheumatoid arthritis (RA) patients.
    The outcome of interest is the probability of being diagnosed with anemia within one year of RA diagnosis
    (also known as risk of anemia),
    and estimation is required in various sub-populations (e.g.~individuals over the age of 65).
    The models must be accurate, and additionally, causal inference assumptions should be met.
    We show how application of our evaluation toolkit can assist in model selection and data refinement as demonstrated below.
    We identified 121,915 individuals diagnosed with RA between 2012 and 2016.
    11 treatment groups were identified in the data, with 5 of them belonging to the TNFA family.
    263 covariates were extracted from the 6 months prior to diagnosis.
    The date of first DMARD prescription following RA diagnosis was defined as the reference date (index-date),
    and an indicator for the diagnosis of anemia in the year following the index date was extracted as the binary outcome.

    In step B of Figure~\ref{fig:eval_pipeline},
    to allow showcasing both propensity and outcome estimation we chose to apply a doubly-robust method
    where the weights provided by the propensity model are used as additional features to the standardization model for outcome prediction,
    using inverse probability weighting (IPW) as the weight estimation method for simplicity.
    In step C we start with the simplest models of logistic regression and proceed to more complex models
    only as indicated by the evaluations as described below.
    Note that the examples below are not intended as a recommendation for which methods and models should be used in general.
    Instead, they are an example for a process to choose the best models for a given dataset.

    \subsection{Evaluation of Propensity Models}
    We next turn to step D in the pipeline in Figure~\ref{fig:eval_pipeline}.
    Propensity models learn the likelihood of being treated
    and are used to emulate randomized controlled trial (RCT) by re-weighting, matching, or stratifying
    \citep{rosenbaum1985constructing}.
    To judge the success of the propensity model, several evaluations can be performed, as described below.
    We will first focus on covariate balancing.
    To emulate an RCT, all covariates must be equally distributed
    (i.e., balanced) across the treatment groups.
    This is often evaluated marginally using a covariate balancing plot,
    in which the difference in distribution between treatment groups is quantified using the absolute standard mean difference (SMD)
    \citep{love2002displaying}.
    A small SMD value indicates good balancing, while a value larger than some threshold is considered unbalanced (or imbalanced).

    \begin{figure}
        \begin{center}
            \includegraphics[width=0.7\columnwidth]{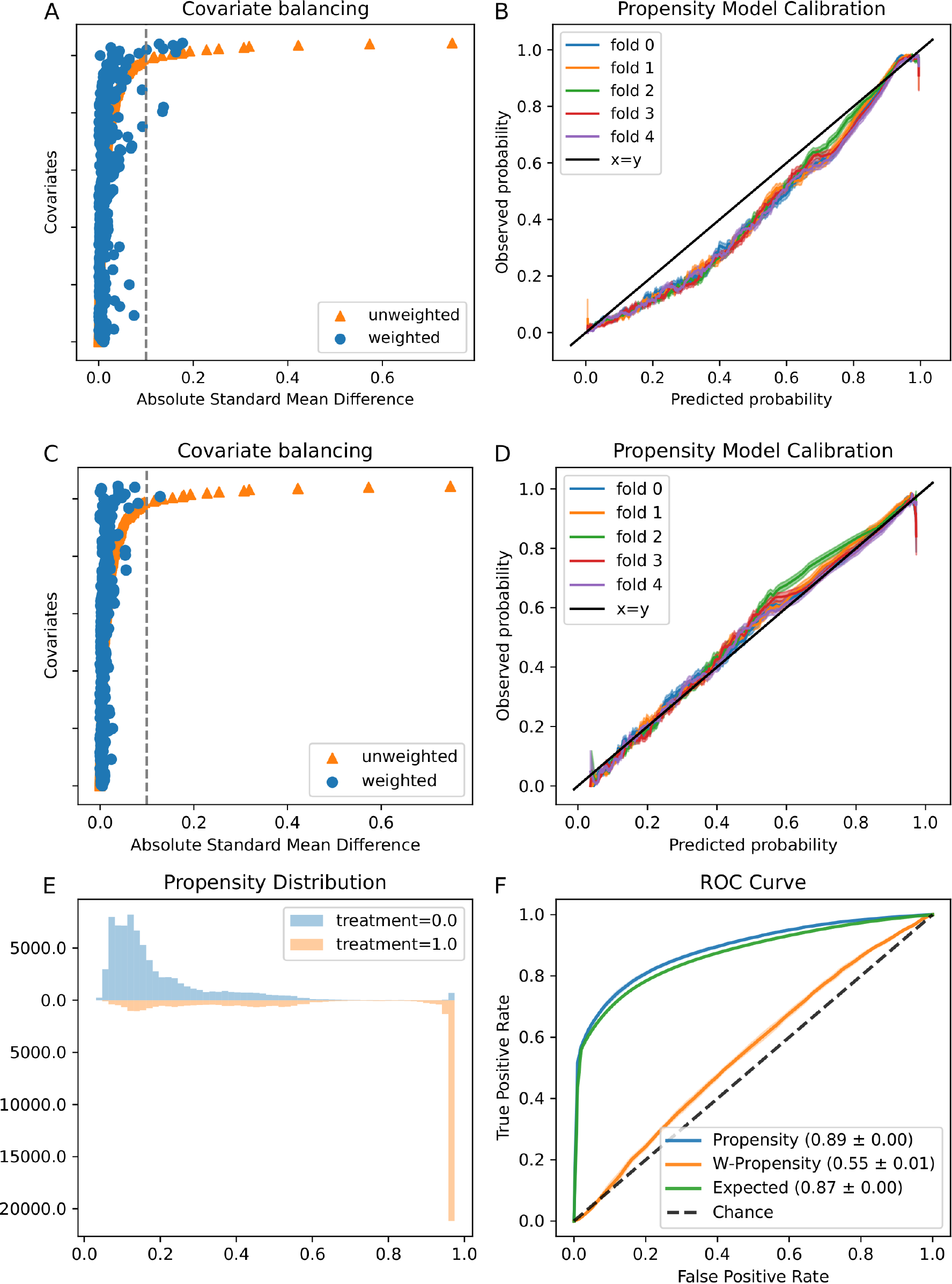}
        \end{center}
        \caption{Performance comparison of IPW using regularized logistic regression (A, B)
        and using calibrated regularized logistic regression (C-F).
        Plots A and C show balancing plots, displaying the absolute SMD of each covariate in the original data
        (orange triangles) and in the weighted data (blue circles), showing that the calibrated model provides better balancing.
        Plots B and D display the calibration of the propensity models for the 5 validation folds,
        comparing the predicted propensity and the observed probability to be treated. As expected,
        the calibrated model provides better calibration.
        Plot E shows the distribution of predicted propensities for treated (orange) and untreated (blue) patients.
        Plot F shows the ROC curves for the propensity model (blue), the expected value (green), and that the weighted propensity (orange).
        Together, the steep ROC curve in plot F and the lack of overlap in the supports of the histograms in plot E
        uncover an apparent violation of positivity in the data.}
        \label{fig:init_eval}
    \end{figure}

    The first evaluation tool we therefore introduce is the covariate balancing plot,
    exhibited in Figure~\ref{fig:init_eval}A,
    where the SMD is displayed as the averaged over the validation folds in a 5-fold cross validation of the data
    (namely, the RA cohort for treatment with a TNFA, as evaluated using IPW with a logistic regression model).
    The blue dots represent the SMD after re-weighting according to the model, while the orange triangles represent the SMD in the original data.
    Many covariates exhibit high imbalance in the original data.
    While most of the covariates exhibit low SMD after re-weighting,
    there are several covariates that exhibit SMD values above the 0.1 threshold
    (which is a customary arbitrary threshold that assumes Gaussian distributions).
    This indicates that the estimated average outcome from this model may be inaccurate, and that another model should be used.

    The use of the estimated propensity is to capture the true probability that an individual is treated.
    When this assumption is not met, the subsequent use of propensity for weighting or matching becomes suspect.
    The next evaluation tool we introduce is the calibration plot, exhibited in Figure~\ref{fig:init_eval}B,
    showing the average predicted propensity on the x axis \vs the observed proportion of treated individuals
    (empirical probability) in the y-axis.
    The plot shows that the predicted propensity values differ significantly from the
    observed propensity to treat for most of the value range.
    Therefore, average outcomes based on these values may be biased.
    Furthermore, even if despite the lack of calibration good balancing is observed,
    since the model does not reliably reflect the propensity, it is unlikely to continue balancing for subsets of the data.

    Since the evaluations above are not satisfactory, we will re-iterate steps A-C of Figure~\ref{fig:eval_pipeline},
    modifying the machine learning model and replacing the logistic regression by a calibrated logistic regression model
    \citep{zadrozny2002transforming}.
    Repeating step D in the pipeline of Figure~\ref{fig:eval_pipeline} as above,
    Figure~\ref{fig:init_eval}C and Figure~\ref{fig:init_eval}D show that this approach improves the SMD
    for most of the covariates while also exhibiting good calibration.
    Thus, the model provides more reliable weight estimations, including those for sub-populations.

    An additional evaluation tool for the propensity model, which is causally oriented,
    assesses whether the data exhibits positivity violations.
    Positivity is a property of the data where every individual has a probability of being assigned any treatment.
    Violation of the positivity invalidates the causal question,
    since it is no longer relevant to ask what might happen to individuals if they change treatment
    when that treatment can never be given to these individuals \citep{hernan2006estimating}.
    Beside this philosophical issue, on a practical level,
    if there is a group of patients that can be characterized in some way for which one treatment assignment
    was not observed, 
    then the machine learning model must extrapolate (rather than interpolate)
    what might happen to this sub-population under treatment change, and so its predictions may not be as reliable.

    Figure~\ref{fig:init_eval}E shows the propensity distribution plot for the analysis.
    We see that many of the treated patients are predicted to have a propensity of 1.
    This is an indication that the model
    was consistently able to identify a volume in covariate space where virtually all patients receive the same treatment assignment,
    violating the positivity assumption.
    In general, violation of positivity is seen in this plot when there is an area in which
    a single treatment value is predicted, as is the case in the far right of the graph.
    In most cases, the way to solve this problem is to identify the defining characteristic
    for this sub-population so that they can be excluded from the analysis.

    In our example, the covariate describing this sub-population was found to be treatment with TNFA before RA diagnosis.
    We then discovered that patients who received any DMARD prior to RA diagnosis
    were almost never given another treatment upon diagnosis.
    Almost no examples exist in the data to allow learning what might happen if these patients received another treatment upon diagnosis;
    therefore, no prediction can be made for what might happen under other treatment conditions.
    Thus, we return to step A in the pipeline of Figure~\ref{fig:eval_pipeline} to re-define the causal question and the data.
    Specifically, we excluded any RA-diagnosed patient who received any DMARD in the year prior to RA diagnosis, 
    reducing the cohort to 92,496 patients.
    The causal question was accordingly updated to
    estimating the effect of treatment with a TNF-alpha antagonist (TNFA)
    compared to treatment with other DMARD,
    for recently-diagnosed rheumatoid arthritis (RA) \emph{DMARD-\naive} patients.

    Note that observing apparent violations of positivity only in the training data (i.e.~not in the validation data) 
    may indicate an artifact of overfitting.
    It should also be noted that violation of positivity can cause a model to calibrate poorly and/or fail to provide good balancing.
    Consequently, violations of positivity should be addressed prior to attempts to improve balancing or calibration.

    \begin{figure}[t]
        \begin{center}
            \includegraphics[width=0.8\columnwidth]{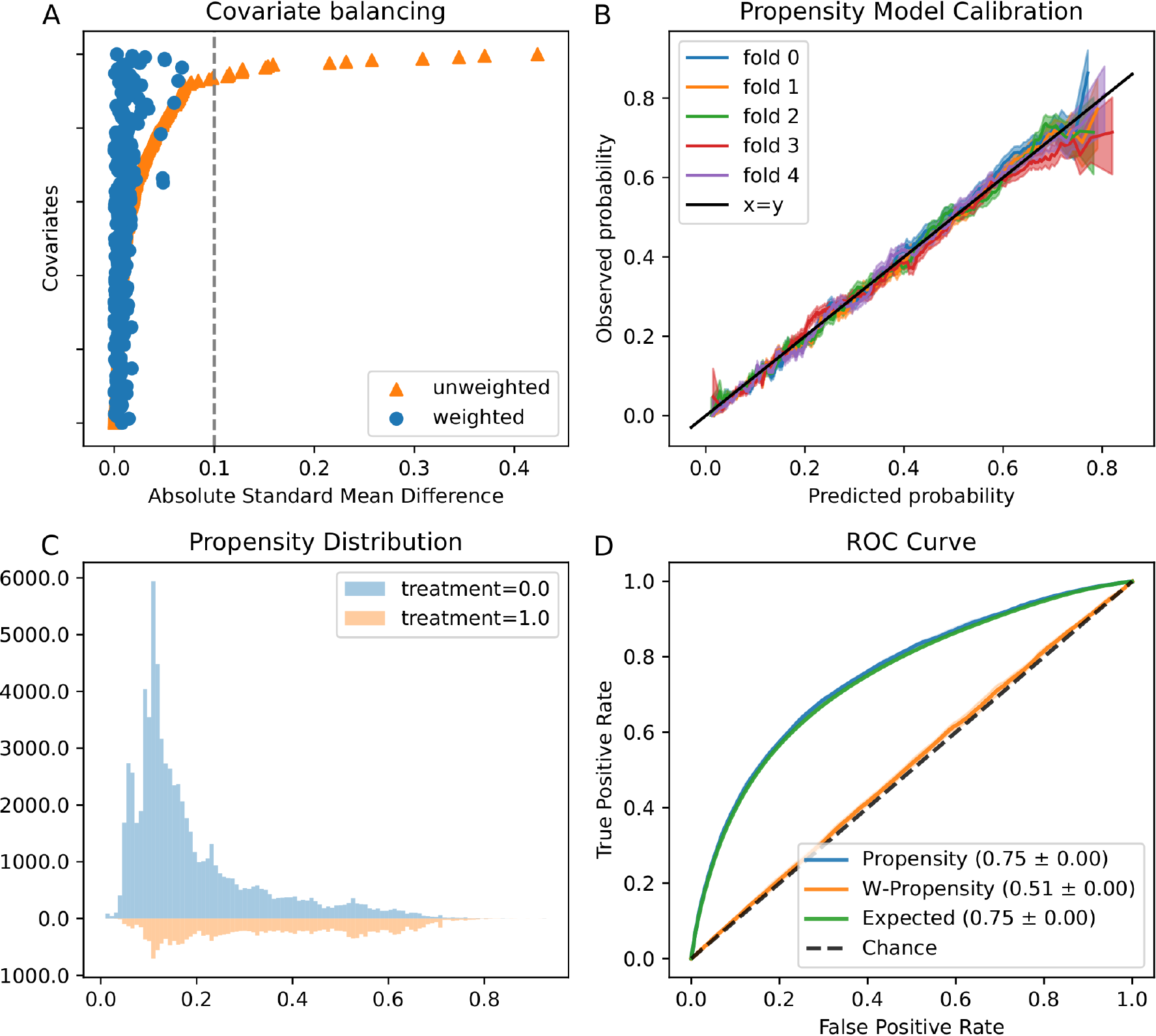}
        \end{center}
        \caption{Performance of a propensity model using calibrated logistic regression
        after correcting positivity violation shown in Figure~\ref{fig:init_eval}.
        Plots A, B, C, D correspond to plots C, D, E, F in Figure~\ref{fig:init_eval}, respectively.
        All the evaluations exhibit better performance.
        Specifically, A) All covariates exhibit good balancing below the threshold of 0.1;
        B) The calibration curves are closer to the desired line presented by the diagonal;
        C) No violation in positivity is observed;
        and D) the ROC plots exhibit the desired behavior, where no steep (or flat) areas appear in the curves,
        the propensity curve (blue) is close to the expected curve(green),
        and the weighted curve (orange) is close to chance prediction.}
        \label{fig:pos_corrected}
    \end{figure}

    Leaving steps B and C in the pipeline of Figure~\ref{fig:eval_pipeline} unchanged,
    Figure~\ref{fig:pos_corrected} presents the covariate balancing plot (A),
    the calibration plot (B) and the propensity distribution (C) for the same model as above after limiting to
    treatment-\naive patients.
    By correcting the positivity problem, the model now achieves better balancing,
    better calibration, and a propensity distribution that does not hint at violations of positivity.

    A final evaluation tool for the performance of the propensity model is obtained using ROC curves,
    which are a standard way of estimating the performance of categorical predictions.
    Usually, a desired result is the steepest curve possible (maximum TPR for minimal FPR).
    However, this is not the case for propensity models.
    Either vertical or horizontal lines (as seen in Figure~\ref{fig:init_eval}F)
    suggest that there are one or more covariates that are highly discriminative between the treated and untreated groups,
    indicating a potential violation of positivity.
    Conversely, AUC values close to 0.5 do not necessarily indicate an inaccurate model.
    For example, modeling the treatment assignment for a proper RCT should have an AUC of 0.5. Empirically,
    AUC values between 0.7 and 0.8, like the values observed in Figure~\ref{fig:pos_corrected}D,
    exhibited the best downstream performance for reliable effect and outcome estimation.

    A novel use for ROC curves, which is only relevant in the context of propensity estimation,
    is as a test for the consistency of the propensity model.
    This curve is obtained by re-weighting the standard ROC curve using the weights drawn from the propensity model
    (i.e., the same weights used for the covariate balancing plots above).
    A model that accurately predicts the propensity to treat should be very close to the curve that would arise by a chance assignment
    (i.e., with an AUC of 0.5).
    This curve is termed the "weighted" ROC\@.
    Figure~\ref{fig:pos_corrected}D shows the weighted ROC curve for the model above,
    exhibiting a weighted AUC of 0.47 with an error that includes 0.5, as required.

    The analyses above indicate that the propensity model that was derived provides reliable and reproducible results,
    and that these results should also be reliable for various sub-populations of the data.

    \subsection*{Evaluation of Outcome Models}
    In many contexts, such as in precision medicine, prediction of average outcomes and average effects may be insufficient.
    In this case, it is necessary to construct a model that predicts the outcome directly,
    which is a more standard problem in machine learning.
    It is therefore important to evaluate the performance of outcome models as well.
    In the context of causal inference, when the standard assumptions of exchangeability and consistency hold
    we can assume that $Y|A=Y^A$.

    Evaluating the performance of outcome models is more straightforward than evaluating the performance of propensity and weighting models,
    and some of the evaluations below are also available using other tools \citep{reps2018design}.
    As mentioned, to evaluate the causal performance of a model we suffice by examining how well the model
    can predict the observed outcomes given the observed treatment.
    This can be justified by the assumptions of exchangeability and consistency,
    namely, that the probability of the outcome for each treatment is independent of the observed treatment assignment.
    Under these assumptions, the expected value of the outcome given treatment is a good estimator of the counter-factual outcome.
    Still, we stress that these evaluations can only detect problems with the model and can never directly evaluate
    the model's performance on effect estimation.

    It was shown that doubly robust methods that estimate both the propensity
    and the outcome can have tighter bounds on their biases compared to each individual estimation \citep{funk2011doubly}.
    We therefore continue the example using a doubly-robust method where the inverse of the propensity
    is added as an additional feature to the machine learning method
    (at step D in the pipeline of Figure~\ref{fig:eval_pipeline}).
    Since the observed outcome of a patient being diagnosed with anemia is binary,
    some of the evaluation plots that were used for propensity can be used,
    namely, the calibration plot and the ROC curves.
    Unlike for propensity, however, it is desirable to have an AUC as large as possible.
    As in usual prediction modeling, large differences in the curves between folds
    may uncover unequal data distribution among folds.
    This, in turn, can skew performance assessment and model selection.
    Additionally, precision-recall curves may be used instead of ROC curves when classes are imbalanced
    (e.g., rare outcomes, like heart failure), since rare events tend to lead to high rates of true-negatives,
    using them, as a ROC curve does, may lead to an overly-optimistic view of the model \citep{fawcett2006introduction}.

    For continuous outcomes, where ROC curves and calibration plots are not applicable,
    accuracy plots are a standard method to evaluate the accuracy of the model's predictions.
    They compare the observed outcome \vs the predicted outcome given the observed treatment
    (i.e., the predicted factual value), where an r-squared estimation can be made.
    Alternatively, a residual plot showing the difference between the predicted and the observed value can be used.
    While there may be multiple reasons why a model fails to provide accurate predictions
    (e.g.,~insufficient covariates, under-specified model, over-fitted model, etc.), this plot,
    especially when applied to both training and validation parts of the cross-validation,
    can help distinguish some of these issues.
    The separation over treatment groups can also allow recognition of class imbalance in the training data.

    We now turn to an outcome evaluation tool that is specific to causal inference, namely ignorability evaluation.
    Ignorability states that the potential outcome of any treatment is conditionally independent 
    of the observed treatment assignment given the covariates.
    This requires that the joint distribution of the potential outcomes are identical in both treatment groups 
    (and hence independent of treatment assignment).
    Ideally, if the true counterfactual outcomes were known, they could be plotted to help identify violations of ignorability.
    Since all potential outcomes cannot be observed,
    we plot the \emph{predictions} of the potential outcomes against each other.
    Violations of ignorability may seem counter-intuitive because they imply that the choice of treatment assignment itself changes the potential outcome,
    which is not possible.
    However, a lack of overlap in the predicted counterfactual outcome distributions may indicate:
    a) that the model is not internally consistent as a causal inference model;
    or b) that there are unmeasured confounders, or that the model disregarded a measured confounder.
    In either case, such a violation of ignorability indicates that outcome predictions from the model are not reliable.

    \begin{figure}[t]
        \begin{center}
            \includegraphics[width=0.8\columnwidth]{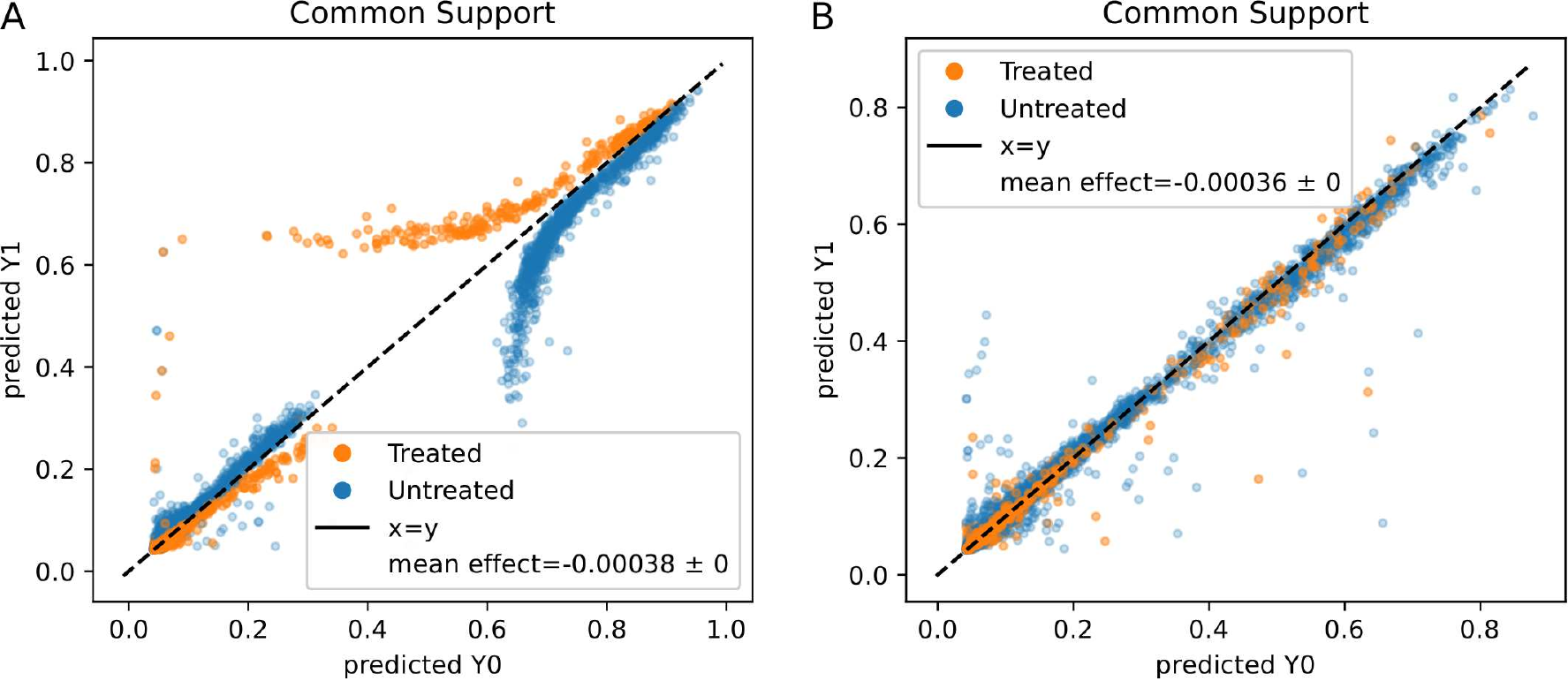}
        \end{center}
        \caption{Predicted outcome plots for the probability of anemia for RA patients treated with TNFAB,
        when estimated using a random forest classifier as part of a doubly-robust model.
        A) The predicted probability with treatment on the y-axis \vs the predicted probability 
        without treatment for patients who were treated (blue)
        and patients who were not treated (orange) for the training data.
        The plot exhibits a violation of ignorability, namely, areas where a single treatment group appears,
        suggesting that it should not be used for causal inference.
        B) The same evaluation on the validation data does not exhibit violation of ignorability but exhibits very small predicted effect.
        }
        \label{fig:random_forest}
    \end{figure}

    Figure~\ref{fig:random_forest}A shows the predicted counter-factual plot for predicting the risk of anemia in the RA cohort
    when the patients are treated with TNFA for the training folds of the same 5-fold cross validation as above,
    when the outcome is predicted using a random forest classifier with 500 trees.
    The plot pattern shows many areas where a single treatment group is evident.
    This indicates that the model predicts some violation of ignorability (or model inconsistency).
    Hence, this model cannot be used to predict the causal effect of treatment and is not reliable for individual outcome prediction.
    Note that this evaluation must be performed on the training folds in order to observe this behavior.

    A likely reason for this behavior can be derived by noting that the distributions of the two treatment groups
    are almost inverted along the $x=y$ line.
    Since random forest models are notorious for over-estimating training performance,
    the factual predictions on the training data are done using out-of-bag estimation.
    However, out-of-bag estimation cannot be used when predicting the counter-factual outcomes,
    since the counter-factual vector was not in the training data.
    On the other hand,
    since the only thing that changes between the factual and counter-factual prediction is the treatment,
    the input vector to the prediction model is extremely similar to the vector that was given during training.
    As a result, the counter-factual predictions are over-fitted to the observed factual outcomes.
    A more detailed derivation of this conclusion appears in \nameref{appendix_B}.

    This analysis is a general one and shows that without out-of-bag predictions for the counter-factual treatments,
    models with inherent high over-fitting and small effect for change of a single covariate (such as random forests)
    cannot be used to obtain consistent causal effects models.
    As a result, the current implementation of random forests in scikit-learn cannot be used
    without significant modifications to predict the effect or outcome on the training data.
    Figure~\ref{fig:random_forest}B shows the same plot for the validation folds, where no violation in ignorability is detected.
    However, since changing a single covariate causes little change to the model,
    such models usually predict very small effects.

    Besides ignorability, the predicted counterfactual outcome plots can help evaluate
    the overall structure of the predicted effect and help identify populations of interest.
    Specifically, this visualization allows us to see if any predicted effect exists,
    whether the predicted effect is consistent across the population,
    or if there are specific subsets of the data for which the effect is predicted
    to be significantly larger or smaller than the rest of the data.
    Additional insights can be gained from this visualization depending on the structure of the distribution.
    For example, a very narrow distribution along some function hints at low-dimensionality of the problem
    because only a few covariates (or summary functions) influence the outcome prediction,
    while a wider and more complex distribution indicates multiple covariates affecting the problem.

    \subsection{Performance on Subsets of the Data}
    An additional use for evaluation toolkit is for understanding model performance on subsets of the data.
    In cross-validation schemes, the model is trained on a training fold and then tested on a validation fold;
    however, for subsets of the data the model can be trained on the whole data and then tested on a subset.
    A model that performed well on all evaluation tasks may still fail on specific subsets of the data,
    especially when the distributions of covariates within the subset differ from the complete dataset.

    \begin{figure}[t]
        \begin{center}
            \includegraphics[width=0.8\columnwidth]{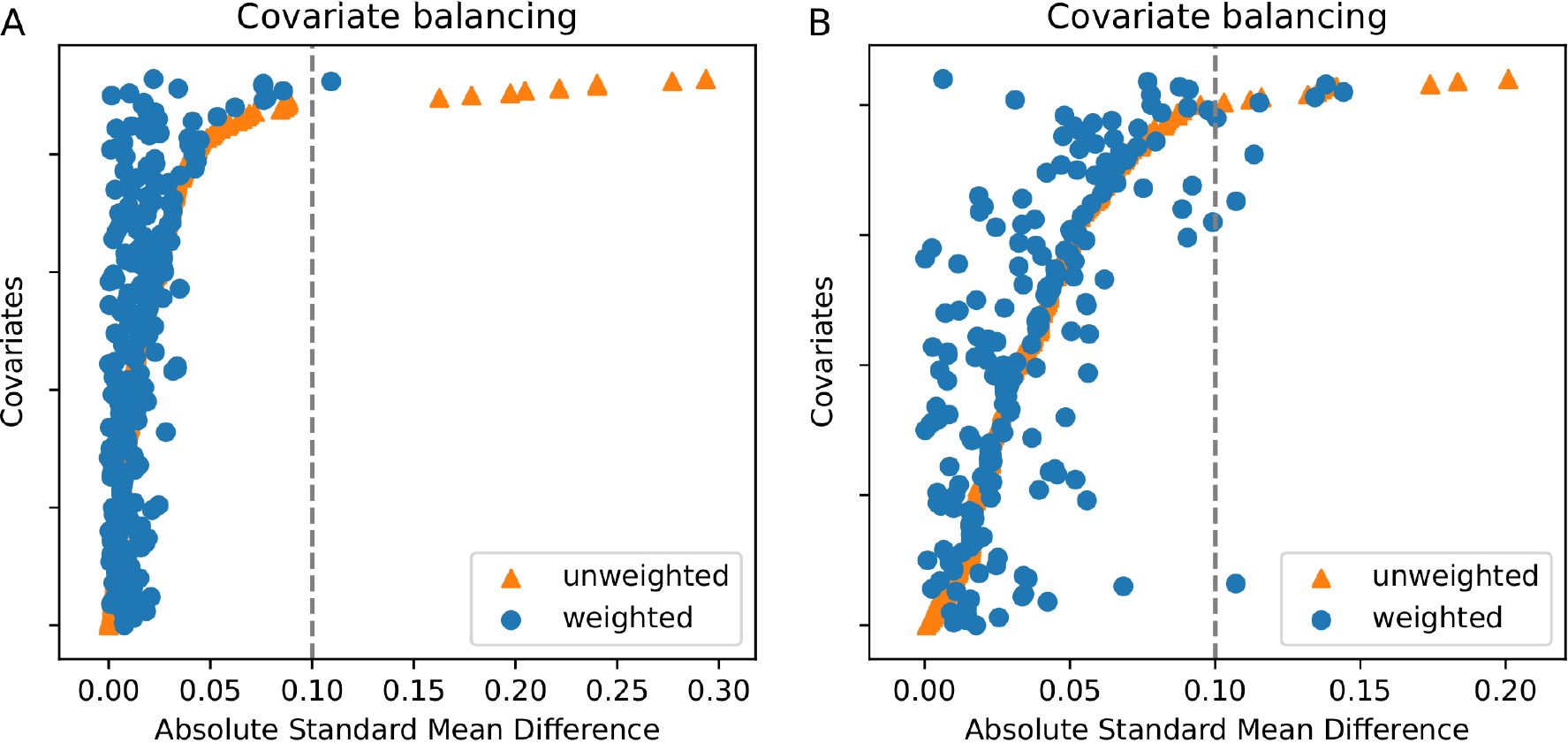}
        \end{center}
        \caption{Balancing evaluation of subsets of the data after training on the complete data,
        displaying the absolute standard mean difference of each feature in the original data (orange triangles)
        and in the weighted data (blue circles).
        A) The plot for all RA patients above the age of 65, which are approximately 28\% of the data, exhibits good balancing.
        B) The plot for RA patients above the age of 75 who also have cardiovascular disease before treatment initiation,
        consisting of about 4\% of the data, exhibits degraded balancing, but within acceptable levels,
        indicating that the model can be used to predict effect in these sub-populations.
        }
        \label{fig:subset}
    \end{figure}

    As an example to how the evaluation may change when evaluating on subsets of the data,
    Figure~\ref{fig:subset}A shows the balancing plot for the same analysis as above using the same models
    (trained on the complete data), but applied to a subset of patients above the age of 65
    (25,480 patients, approximately 28\% of the population).
    Calibration is somewhat worse than was observed for the complete data (Figure~\ref{fig:pos_corrected}A), 
    and a few covariates exhibit larger SMD than before,
    with one covariate exhibiting SMD slightly above the threshold of 0.1.
    Significant degradation in performance is only observed when further restricting the analysis
    to patients over the age of 75 with a history of cardio-vascular disease (4,111 patients, 4\%).
    This is shown in the balancing plot in Figure~\ref{fig:subset}B, where multiple covariates are above the 0.1 threshold.
    Still, in many contexts even these results are considered acceptable.

    Once significant degradation in performance is discovered in a sub-population,
    the action to be followed depends on the context.
    In some contexts, the degradation in performance will only be reported,
    while in others the models can be re-trained for the sub-population (with a risk for overfitting).
    In other cases, the models for the complete dataset may be revised.

    For completeness we report the resulting predicted effect, and how it changes in these sub-populations.
    Since this is only an example we will not discuss the implications of these results.
    The provided models predict a mild decrease of 0.84\% in the risk of anemia for RA patients with TNFA compared to other DMARDS.
    In the population over the age of 65 the effect was slightly more pronounced for a decrease of 1.1\%,
    while in the population over the age of 75 with history of CVD the effect was even larger with a decrease of 1.5\%.
    While these effects are relatively small, they are larger than the standard deviation that was observed in cross-validation,
    which was less than 0.1\%.

    \section{Discussion}
    We propose a causal inference evaluation toolkit to assess the performance of causal algorithms
    by evaluating their machine learning core.
    We do this by evaluating performance against the factual data,
    since for casual models to be truly correct they must at least be correct on the observed data.
    We stress that the logic is unidirectional:
    good model performance on the factual data does not guarantee good counterfactual modeling,
    especially when there are unmeasured confounders.
    However, poor model performance on the factual data does imply poor model performance of the entire causal model.

    This paradigm shift - from predicting effect to predicting propensity and outcome - may be more onerous.
    However, it allows moving from predictions of unknowable quantities (effect)
    to predictions of verifiable and testable quantities.
    It allows evaluating the model for generalizability and its performance on the available data,
    guiding the choice of the most appropriate methods and models for the data being analyzed.
    Furthermore, it also suggests an iterative process of refining the data extraction pipeline
    (i.e.\ changing the inclusion and exclusion criteria, and modifying the covariates available for the models ),
    and at the least provides insight into which questions can be reliably answered and which cannot.
    A proposed guideline is provided below for how such a process can be performed.

    Evaluating propensity models can be highly informative for understanding biases in the data
    and helping discover positivity violations.
    It may therefore be useful to construct a propensity model and evaluate how well it performs,
    even when the main task is to evaluate an outcome directly and there is no plan to use the propensity in outcome estimation.
    Specifically, the propensity model evaluation can be used to detect that a change in the causal question is required,
    or that additional covariates are needed, as explained below.

    Violations of positivity indicate that inclusion and exclusion criteria need to be changed in the data,
    or that a covariate that holds information about the outcome was erroneously introduced.
    Similarly, issues with covariate balancing may indicate that more covariates or a more expressive model is needed.
    Increasing the complexity of the model in this way may lead to overfitting,
    which can be detected by comparing performance between training and validation.
    Lack of calibration may indicate insufficient sample size.
    However, it may also indicate that the model is unable of capturing the underlying structure in the data;
    or the existence of model overfitting.
    If the covariate balancing and AUC are sufficiently good,
    then calibration only reflects on the reusability of the model for other tasks
    (e.g., comparing to different models or applying the results on a subset of the data).
    If a propensity model cannot be reliably constructed, it may be ill-advised to use a doubly-robust method,
    since it inherently utilizes the propensity estimations.

    When evaluating outcome models, the precision evaluations should be used for model selection and covariate selection
    while evaluating how this affects performance on both training and validation folds to avoid overfitting.
    To evaluate the effect and ignorability, however, it is normally relevant to examine training performance,
    since effect predictions are traditionally reported on the same data on which models were constructed.
    Finally, if a reliable outcome model cannot be constructed while a good propensity model is available,
    then it is still possible to use the propensity model to evaluate the effect of treatment in various subsets of the data.

    As with all machine learning evaluations, when applying the above guidelines,
    it is important to remember that making decisions based on the validation folds leads to
    training contamination of the validation samples and may therefore lead to overly optimistic results.
    Similarly, good generalizability on a cross-validation toolkit
    does not guarantee generalizability in other datasets due to batch effects, specific data biases, etc.
    The evaluations and plots described here are only a handful of the evaluations and summaries that can be performed,
    but we believe they are the most important.

    While not mentioned explicitly, our evaluation toolkit is also useful for evaluating matching methods \citep{stuart2010matching}.
    Matching allows the estimation of counterfactual outcomes and is therefore applicable for outcome evaluation.
    It can also be considered an integer-weighting algorithm and can therefore be evaluated using the weighting evaluations.
    Furthermore, specifically, propensity matching requires a propensity model and therefore lends itself to propensity evaluation.

    Our evaluation toolkit may be of particular importance for personalized medicine,
    where generalizability, accuracy, and robustness are imperative,
    but where individual effect is unobservable \citep{shalit2016estimating}.
    Prediction evaluations for personalized models can only be performed on a population-basis,
    as described here for sub-population.
    Only when a model displays sufficient accuracy for a specific sub-population can it be used for precision medicine.

    We firmly believe that causal inference analysis can be used to estimate unbiased measures
    and help guide evidence-based decision making in healthcare.
    However, for both care-givers and care-takers to trust the models in such sensitive matters,
    there's a strong need to validate, understand and judge their robustness and biases.
    We hope this work is another step in that important direction.

%    \printbibliography
    \bibliography{causality}

%    \clearpage
%    \newpage

    \section*{Appendix A: Detailed Methods}
    \label{appendix_A}
    This section presents our evaluation framework using numerical performance metrics and plots.
    The calculations are done using cross-validation to identify over- and under-fitting,
    reduce optimism bias and evaluate the model's generalizability to other similar patient cohorts
    \citep{friedman2001elements}.
    Evaluations can also be stratified by treatment type, allowing assessment of each potential outcome model separately.
    The following describes our framework methodology.

    \subsection*{Code Availability}
    The code for our evaluation framework was written in Python 3.6 and is designed as a wrapper
    for models in a format compliant with scikit-learn \citep{scikit-learn}.
    The evaluation plots were created using matplotlib \citep{Hunter2007} and seaborn \citep{waskom2014seaborn},
    and most of the evaluation metrics are as defined by scikit-learn.
    To request usage of the code please contact the authors.

    \subsection*{Covariate Balancing Plot}
    Many studies report the mean and standard deviation of covariates in each treatment group,
    as well as the standardized mean difference (SMD) of each covariate between the groups.
    The SMD is used as a measure of the marginal similarity between the covariate distributions in each of the groups.
    Unlike other measures of similarity, the SMD is independent of the covariate's scale,
    thus allowing similarity judgments possible across all covariates.
    To allow easy interpretation of the SMD, a plot, commonly known as a Love plot \citep{love2002displaying},
    is provided that shows the SMD of each covariate with and without weighting.
    The covariates are ordered along the y-axis, sorted by the SMD in the original (uncorrected) data,
    and a threshold value is plotted using a dotted line (by default at SMD=0.1, as is common in the literature).
    Good balancing is exhibited when the weighted SMD for all covariates is smaller than the threshold,
    as observed in Figure~\ref{fig:pos_corrected}A.

    Other measures of similarity between the distributions can also be displayed,
    including the symmetrized Kullback-Leibler divergence, the p-value of a Kolmogorov-Smirnov test,
    or the earth-mover distance.
    These measures, however, share assumptions that may not be relevant in the context of causal inference
    \citep{ali2015reporting, linden2015graphical}, or are scale dependent.

    \subsection*{Probability Calibration}
    In machine learning, calibration is relevant in classification problems to assess model fit.
    It is especially relevant when comparing multiple models against each other.
    Here we apply it to assess the fitness of causal inference models,
    where it can be applied to either propensity or outcome modeling.
    Specifically, calibration is used to estimate whether model scores
    (i.e., the continuous output from the classification model prior to deciding on the predicted class assignment)
    can be used as probabilities.
    The predicted scores are evaluated either using a running average, also known as a sliding window, or by binning.
    The plot, also called a reliability diagram, displays the observed frequency of the event in the bin
    on the y-axis \vs the mean predicted probability on the x-axis within the sliding window or bin
    \citep{zadrozny2002transforming} (see Figures~\ref{fig:init_eval}B, \ref{fig:init_eval}D, \ref{fig:pos_corrected}B).
    When using cross-validation, each fold of the data is represented by a different color line
    with the a shaded area surrounding it, representing the approximate 95\% confidence interval.
    This interval is calculated as the values, $r$, for which $r$ plus/minus the standard deviation of the bin
    (using binomial distribution) would equal the observed proportion, $p$,
    namely $r \pm \sqrt{\frac{r(1-r)} {N}} =p$, where $N$ is the number of samples in the bin.
    In order to compare to the desired result, the line for $x=y$ is also provided.

    \subsection*{Propensity Distribution by Treatment Assignment}
    Positivity in causal inference is the assumption that there is
    some likelihood for any patient to receive or not receive treatment.
    A violation of this assumption negates causal analysis,
    since it means that for some set of covariates the treatment is known
    and that other treatment options are never observed.
    This suggests that there is probably a reason (either regulatory or practical)
    that disallows the unobserved treatment assignment.
    Regardless, since one of the treatment groups is not observed,
    a model trained on such data is unlikely to be able to reliably predict
    the values for that treatment assignment over that subspace.

    The positivity assumption must be met in the multidimensional space of all the covariates,
    which is not feasible to visualize.
    However, it was shown \citep{rosenbaum1983assessing}
    that it is theoretically sufficient to check that the positivity is met in the propensity space.
    Therefore, to evaluate whether the positivity assumption is violated a plot is provided that displays
    the distribution of propensity scores (or weights) for the different treatment groups.
    Each treatment assignment is represented by a different color
    (see Figures~\ref{fig:init_eval}E, \ref{fig:pos_corrected}C).
    Violations of positivity are exhibited by areas for which a single treatment assignment is observed.
    The distribution can be plotted using a histogram,
    the probability density function (PDF) or the cumulative density function (CDF).
    The tradeoff is that histograms and PDFs are more easily interpretable
    but plotting them requires some approximation - either by assigning to bins (histogram) or by using kernel
    density estimations (PDF), both can lead to some artifacts depending on the parameters used.
    In contrast, empirical CDFs can be plotted with no artifacts at all but can be less intuitive to interpret.
    Ideally, the support of all distributions should coincide between the groups.
    To improve clarity of overlap between groups, when the treatment is binary and PDF is used,
    the PDF for the treated samples is reflected around the x-axis.

    Note that propensity estimation can be viewed as a dimensionality reduction (to a single dimension).
    This may lead to different sub-spaces being mapped to similar propensity values,
    thus obscuring a violation in positivity in one of them.
    We therefore stress that no observed violation of positivity in the propensity space
    does not guarantee no violation of positivity in the multidimensional space,
    while an observed violation in positivity in the propensity space
    does indicate a real violation of the positivity assumption.

    \subsection*{Prediction Accuracy Plots}
    To judge the accuracy of a model, whether for propensity or outcome,
    we apply several standard metrics with visualizations.
    The choice of metric and visualization is determined by the type of variable being modeled,
    dichotomous or continuous.
    We discuss each of these options below.

    For dichotomous outcomes, the ROC curve is commonly used in statistics to plot a model's prediction against a
    known truth as the discrimination threshold is varied.
    The true positive rate (TPR, aka sensitivity or recall) is plotted on the y-axis \vs the false positive rate
    (FPR, aka 1-specificity) on the x-axis (see Figure~\ref{fig:pos_corrected}D).
    When using cross-validation, the TPRs are first linearly interpolated over the range of zero to one.
    This allows to align the various folds as a function of the FPR in order to
    calculate average and standard deviation across the different folds at each point along the x-axis.
    The average is then plotted as the curve, and the standard deviation surrounds it.
    Optionally, a dedicated fainter curve for each fold can be plotted instead of the shaded area,
    but this may lead to clutter.
    Finally, the area under the ROC curve (AUC) is calculated for each fold separately
    and the average AUC is presented in the legend $\pm$ 1 standard deviation
    (the AUC for the average plot is not calculated since it is not a good statistical measure).

    In general, $N_A$ curves are drawn, where $N_A$ is the number of possible treatments.
    For each treatment $a \in A$, we evaluate the predicted $\{\hat{Y^a_i} | A_i = a\}$ against the true 
    $\{Y_i |A_i= a\}$ and each such comparison has its own curve in the plot.

    All curves, including the propensity and outcome model (when relevant, e.g.\ for doubly robust models),
    can be displayed on the same plot.
    A line for $x=y$, depicting chance prediction, is provided for comparison.

    For propensity models, two additional novel curves can be drawn: a) propensity balancing;
    and b) the expected propensity ROC:
    a) To estimate that the propensity value are balanced across treatment groups, the ROC curve can
    be drawn using the assigned sample weights.
    Similar to a randomized controlled trial (RCT), the weighted
    distribution of propensities in each treatment group should be identical, and so this plot should be as close as
    possible to the random assignment.
    This is similar to a two-sample discrimination test on the weighted samples.
    Significant deviation from the diagonal indicate a deviation from the assumption that the weighting can emulate
    an RCT\@.
    b) The consistency of the propensity model can be evaluated by considering the expected ROC curve
    assuming that the individual propensity predictions are correct.
    In this scenario, the probability of to be treated is their propensity.
    Therefore, each patient contributes their propensity to the true-positives and the
    complement of their propensity to the false positives.
    This curve can be used as a comparison for the observed propensity ROC curve.
    Significant deviations from this curve indicate that the model provides values that are not
    consistent with its interpretation as propensity.

    For continuous outcomes, ROC curves cannot be utilized to estimate prediction accuracy,
    therefore, a scatter plot of the predicted values \vs the true values is used instead.
    Dots in the plot are colored by treatment to visualize if the goodness of prediction is similar across groups.
    In order to compare to the desired result, the line for $x=y$ is provided.
    The r-squared for each of the treatment groups is provided in the legend $\pm$ 1 standard deviation.
    Since these are standard plots no examples are provided in the manuscript.

    To evaluate bias, a residual plot can be used instead showing the residual
    (i.e., the difference between the predicted and observed values) can be plotted in the y-axis.

    \subsection*{Common Support Proxy}
    The ignorability assumption is key in causal inference methods because it is related to the
    underlying strength of clinical trial design.
    Random assignment in clinical trials ensures that all measured and unmeasured confounders
    are equally distributed between treatment groups.
    Therefore, any difference in outcome can be attributed to the treatment.
    The strong ignorability assumption in causal inference states that if all measured confounders
    are equally distributed between the treatment groups then
    we can ignore any unmeasured confounders and attribute any difference in outcome to treatment.

    To test the ignorability assumption, we again draw from machine learning methods and created a plot inspired by
    \citep{hill2013assessing}.
    In this plot the predicted outcome of an intervention (say, control group)
    on the x-axis \vs the predicted outcome of another intervention (say, the treatment group)
    on the y-axis (see Figures~\ref{fig:random_forest}A, \ref{fig:random_forest}B).
    The line for $x=y$ is added for reference.
    In this scatter plot samples that were treated are displayed in a different color than points
    representing samples that were not treated.
    Violation of ignorability are observed when areas of the plot contain points from a single treatment group.

    Note that ignorability requires an identical distribution between treatment groups
    in the hyper-dimensional space of all the covariates,
    while this plot projects these distributions into a two-dimensional space.
    This plot therefore does not guarantee that ignorability is not violated.

    \subsection*{Detailed Metrics Files}
    In addition to the visualizations described above,
    we also recommend creating a comprehensive table with scores in csv format to allow
    use-case specific evaluations and visualizations.
    More importantly, the summary evaluations allow to automatically select models
    based on their performance on a large scale, e.g.\,
    when there are many outcomes and many possible treatment assignments.

    Each line in the csv file corresponds to the metric evaluation for a treatment (TX) \vs 
    outcome (O) pair in a specific fold and a specific train or validation phase (phase)
    and for a specific treatment group (stratum).
    Therefore, each line will have the format: TX, O, phase, fold number, and stratum,
    followed by the relevant metrics (e.g.\ root mean squared error (RMSE) for regression,
    area under ROC curve (AUC) for classification, etc.).
    For completeness, an additional special 'stratum' of the overall population pooling of the treatment groups is
    also evaluated.
    For propensity models the outcome column is omitted.

    The metrics we recommend to be evaluated (when relevant) are:
    accuracy, precision, recall, f1 score, ROC AUC, hinge loss, Matthews correlation coefficient, 0/1 loss,
    Brier score loss, the confusion matrix, explained variance, mean absolute error, mean squared error,
    mean squared log error, median absolute error, and $r^2$ score.
    All these scores are as defined by the scikit-learn.metrics package.

    An additional csv file details the SMD for each covariate in each of the folds with and without weights.
    Each line in this file includes the TX, O, phase, the covariate name,
    the SMD values in all training phases, and all the SMD values in all the validation phases.

    \section*{Appendix B: Counterfactual Overfitting in Random Forest Models}
    \label{appendix_B}
    This section explains in more details the observed behavior seen in Figure~\ref{fig:random_forest}A,
    where the predicted outcomes for the two treatment groups present a mirror image along the $x=y$ line.
    Below we explaing why we believe this can be explained by overfitting of the counter-factual predictions
    in the context of random forests.

    Out-of-bag predictions apply only the random trees in the random forest
    that did not use a specific sample in order to predict the outcome for that sample,
    and were shown to be essential in reducing over-fitting in random forest models \citep{liaw2002classification}.
    However, by the nature of this definition this prediction can be easily be constructed during training,
    and very difficult to maintain for prediction.

    Consider an example of two samples, $i$ and $j$, with almost identical covariates $x_i \approx x$ and $x_j \approx x$,
    respectively, where $i$ was treated and $j$ untreated.
    Let us denote the out-of-bag predictions by $b(x)$ and the regular predictions by $f(x)$.
    For sample $i$ the prediction under treatment will be out-of-bag,
    while the prediction under no treatment will be the regular one.
    It will therefore appear in the plot at $(f(x_i, a=0), b(x_i, a=1) \approx (f(x, a=0), b(x, a=1)$.
    Similarly, for sample $j$ the prediction under no treatment will be out-of-bag
    and so it will appear in the plot at $(b(x, a=0), f(x, a=1)$.
    Since for random forests the change in a single variable is inherently small,
    the point for $i$ and $j$ are approximately equal to
    $(f(x, a=0), b(x, a=0)$ and $(b(x, a=0), f(x, a=0)$, respectively, which are mirror images of each other.

    In addition to this general analysis, in the detailed metric files that are produced during the analysis
    it was indeed apparent that the counter-factual predicted values with no treatment
    provided almost perfect performance when compared to the observed outcome on the treated population,
    and vice-versa.
    This performance was much better than the one exhibited by the factual predictions
    (e.g.\ predicting outcome with treatment for the treated population).

\end{document}